\documentclass{article}

\usepackage[dblblindworkshop,final]{neurips_2026}

\usepackage[utf8]{inputenc}
\usepackage[T1]{fontenc}
\usepackage{textcomp}
\usepackage{hyperref}
\usepackage{url}
\usepackage{booktabs}
\usepackage{amsfonts}
\usepackage{amsmath}
\usepackage{amssymb}
\usepackage{nicefrac}
\usepackage{microtype}
\usepackage{xcolor}
\usepackage{graphicx}
\usepackage{array}

\workshoptitle{Continual World Models}

\title{Physical Experience Builds Continual Spatial World Models}

\author{%
  Yuhang Sun \\
  University of Washington \\
  \texttt{sunyhg@uw.edu} \\
}

\begin{document}

\maketitle

\begin{abstract}
World models are increasingly used to support prediction and planning in
embodied agents, but the knowledge is usually established before
deployment and then carried into later encounters. Continual operation
faces a different problem. A familiar environment may change, and the
agent must use previous experience without allowing old expectations to
override current physical evidence. This study asks what repeated
physical experience must contribute for a spatial world model to become
progressively more useful rather than simply more persistent. A
physics-grounded dynamic maze separates recurring spatial change from
the continuous consequences of action. Cross-episode Spatial Change
Memory learns relations among environmental transformations, while
state-action-outcome experience refines predictions of motion.
Structured experience progressively reduces the need for future route
correction and transfers to spatial transitions not encountered during
learning. The advantage disappears when past changes contain no
information about future ones. Continuous navigation further shows that
predictive memory changes behavior when a previous decision becomes
invalid but leaves behavior unchanged when the decision remains correct.
Physical tests expose a complementary limit: correct topology does not
guarantee correct action prediction, and accumulated physical experience
reduces this mismatch within the modeled dynamics. The results
characterize continual world knowledge as experience that improves
future prediction and action beyond stored episodes while remaining
revisable by the physical outcomes it is intended to predict.
\end{abstract}

\section{Introduction}

The ability to predict the consequences of action has shaped model-based
learning from its earliest formulations. Dyna used observed transitions
both to improve a model and to plan with it {[}1{]}, and PILCO later
showed that data-efficient control could be obtained from learned
probabilistic dynamics {[}2{]}. Deep world models extended the same
principle to high-dimensional observations: recurrent latent models,
PlaNet, Dreamer, MuZero, and more recent general-purpose world-model
agents demonstrated that compact predictive states can support planning,
imagined rollouts, and policy learning {[}3-7{]}. These systems
established the practical value of internal predictive models.
Deployment, however, changes the question. A robot does not encounter a
single stationary environment once. It returns to places whose
accessibility, geometry, and dynamics may evolve, so a model that was
adequate yesterday can become incomplete without any failure in the
planner itself.

Spatial navigation makes this problem especially transparent.
Cognitive-map experiments suggested that animals can represent spatial
relations beyond fixed stimulus-response associations {[}8{]}, and
place- and grid-cell studies established neural correlates of location
and movement across experience {[}9,10{]}. Robotics developed analogous
capabilities through probabilistic mapping and localization {[}11,12{]},
structured neural memory {[}13{]}, topological memory {[}14{]},
long-horizon scene memory {[}15{]}, and learned topological navigation
{[}16{]}. A complementary planning literature asked how computation from
earlier problems could be reused after the environment changes. D* Lite
and Lifelong Planning A* retain information from earlier searches
{[}17,18{]}, while Experience Graphs reuse successful motions to bias
subsequent planning {[}19{]}. These contributions show that memory and
previous computation can make navigation more efficient. They also
delimit the question addressed here: route reuse, incremental
replanning, and episodic storage are not by themselves evidence that a
world model has learned how a changing environment behaves.

Continual learning has largely focused on retaining competence while new
data arrive, including regularization, replay, and architectural
mechanisms for limiting catastrophic forgetting {[}23,24{]}. Online
model-based reinforcement learning has begun to address a related
requirement by updating predictive dynamics during continued interaction
{[}25{]}. For spatial world models, the unresolved issue is more
specific. A robot may detect a familiar blockage and replan successfully
on every visit, yet the tenth visit can require nearly the same new
observation and correction as the first. Repeated success can therefore
coexist with little accumulation of reusable spatial knowledge.
Conversely, a large episode archive may help only when the current
situation closely matches a stored one. A stronger test is whether
experience acquired elsewhere changes prediction on a held-out encounter
and reduces the amount of correction needed later. This shifts attention
from memory of previous states to regularities in how spatial states
transform across encounters.

Embodiment adds a second source of incompleteness. A route can be
topologically valid and still fail during execution because rolling
inertia, dissipation, contact, or actuator response changes the
consequence of a command. Physics-aware control and system
identification have long treated these effects as essential to reliable
behavior {[}20,21{]}, and robotic world-model formulations increasingly
emphasize action-conditioned prediction in the physical environment
{[}22{]}. The point is not that navigation has ignored dynamics. Rather,
spatial memory and action dynamics are commonly represented at different
levels of the system: one determines where a transition exists, while
another determines whether the body can realize it from the current
state. The resulting internal map may be spatially correct yet
insufficient to predict the outcome of action.

\begin{figure}[tbp]
  \centering
  \includegraphics[width=0.98\linewidth]{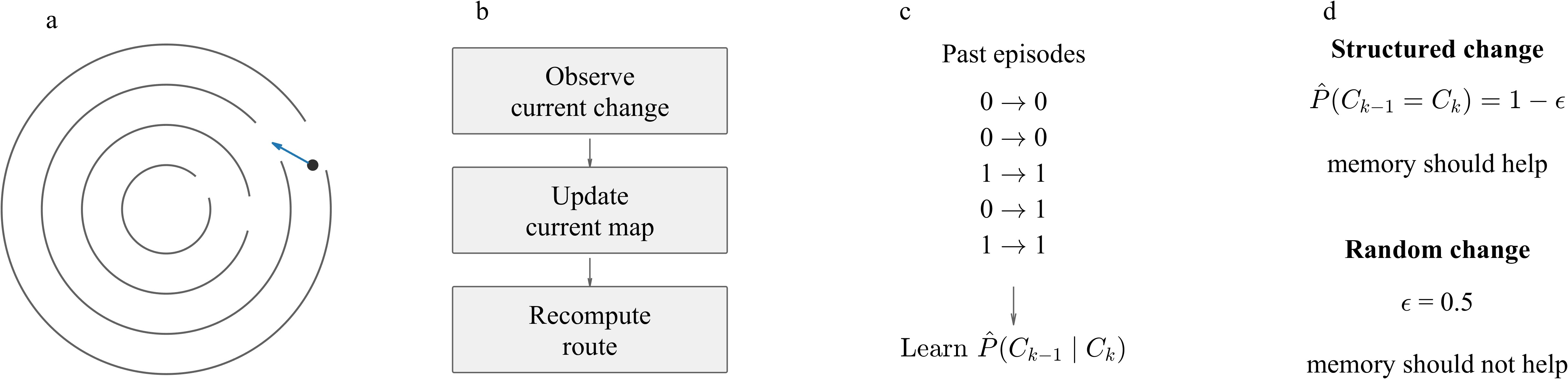}
  \caption{Conceptual organization of the continual spatial world-model experiment. \textbf{(a)} Repeated navigation exposes the agent to related spatial changes across episodes. \textbf{(b)} Reactive updating observes a change, revises the current map, and recomputes the route, so each event is resolved from current evidence. \textbf{(c)} Spatial Change Memory pools transition observations across episodes to learn a relation shared across spatial locations. \textbf{(d)} The falsifiable boundary: structured change should support memory-based prediction, whereas random change ($\varepsilon = 0.5$) should not.}
  \label{fig:concept}
\end{figure}

A deliberately small tilted circular maze is used here to separate these
questions experimentally. Gate states can change between encounters,
while a rolling ball is governed by inertia, dissipation, compliant
contact, and finite actuator lag. Spatial Change Memory (SCM) pools
evidence about recurring relations among changes instead of caching
complete routes. A complementary state-action model accumulates physical
transitions and updates predictions of action consequences. The study
rests on two testable assumptions. First, repeated environmental changes
may contain transferable statistical structure; when that structure is
removed, additional history should not improve prediction. Second, over
the operating range considered here, the dominant physical consequences
of action can be represented by an interpretable low-dimensional
dynamics model. The experiments therefore combine a zero-information
change control, a spatial transition withheld entirely from training,
matched decision-changing and decision-preserving navigation episodes,
and controlled physical-model mismatch. The aim is not to claim a
general solution to lifelong robotics, but to identify when repeated
interaction produces knowledge that is reusable across encounters and
still corrigible by the physical world. Figure~\ref{fig:concept}
summarizes the experimental logic: repeated navigation and reactive
updating (Figure~\ref{fig:concept}(a,b)), cross-episode Spatial Change
Memory (Figure~\ref{fig:concept}(c)), and the structured-versus-random
predictability boundary used as a falsifiable control
(Figure~\ref{fig:concept}(d)).

\section{Continual physics-grounded spatial world model}

\subsection{Physical state, spatial belief, and action prediction}

The maze contains concentric annular regions separated by walls with
discrete gates. Its external spatial state is
\(\mathbf{z}_{t}=\{z_{k,j}(t)\}\), where \(z_{k,j}\in\{0,1\}\)
denotes whether gate \emph{j} on ring \emph{k} is open. The agent
maintains a separate belief \({\widehat{\mathbf{z}}}_{t}\). An
intervention in the external maze therefore does not appear in the
internal model until it is observed or inferred. The continuous state is
\(\mathbf{x}_{t} = [\mathbf{p}_{t},\mathbf{v}_{t},\boldsymbol{\theta}_{t}]\),
containing ball position, velocity, and realized maze tilt.

For a homogeneous solid sphere, rotational inertia enters planar rolling
through \(m_{\text{eff}} = m + I/R^{2} = 7m/5\). The predictor uses

\begin{equation}
\dot{\mathbf{p}} = \mathbf{v},\quad
m_{\text{eff}}\dot{\mathbf{v}} =
m\mathbf{g}_{\parallel}(\boldsymbol{\theta}) - c_{1}\mathbf{v}
- c_{2}\|\mathbf{v}\|\mathbf{v} + \mathbf{F}_{\text{contact}}
  \label{eq:rolling-dynamics}
\end{equation}

where dissipative losses contain linear and speed-dependent terms.
Compliant wall contact is represented by

\begin{equation}
\mathbf{F}_{\text{contact}} =
\sum_{q}\max\left(0,k_{w}\delta_{q} - c_{w}v_{n,q}\right)\mathbf{n}_{q}
  \label{eq:contact-force}
\end{equation}

and the commanded tilt \(\mathbf{u}_{t}\) is realized through

\begin{equation}
\tau_{a}\dot{\boldsymbol{\theta}} =
\operatorname{sat}(\mathbf{u}_{t}) - \boldsymbol{\theta}_{t}
  \label{eq:actuator-response}
\end{equation}

The model retains only physical effects that alter the mapping from
action to outcome in this system: rolling inertia, dissipation, contact,
and finite actuator response. Effective-mass, free-roll,
actuator-response, and contact-resolution checks were completed before
the navigation experiments. Their role is numerical validation rather
than an additional learning task.

\subsection{Accumulating spatial and physical experience}

Structured environmental change is represented by a binary
route-relevant state \(C_{k}\). Across encounters the controlled process
satisfies

\begin{equation}
P(C_{k - 1} = C_{k}) = 1 - \varepsilon,\quad
P(C_{k - 1} \neq C_{k}) = \varepsilon
  \label{eq:change-process}
\end{equation}

Here \emph{k} indexes adjacent spatial transitions/rings, not episode
time, and that \emph{C}\textsubscript{k} is the observed cue used to
predict downstream \emph{C}\textsubscript{k-1}. SCM accumulates
conditional evidence for this relation across equivalent spatial
transitions rather than maintaining an independent memory for each gate.
With a symmetric Beta prior, the posterior predictive probability is
obtained from accumulated compatible and incompatible counts. Sharing
the estimator across locations is essential: evidence collected at one
transition can inform another transition that has not itself appeared in
the training episodes.

The process also supplies an analytical information boundary. With
balanced binary states and no informative cue, the minimum error is
\(1/2\). Conditional on \(C_{k}\), the Bayes error is \(\varepsilon\),
giving Bayes accuracy \(1 - \varepsilon\). The available information is

\begin{equation}
I(C_{k};C_{k - 1}) = 1 - H_{2}(\varepsilon)
  \label{eq:mutual-information}
\end{equation}

At \(\varepsilon = 0.5\), history is statistically independent of the
future state, so additional episodes should not improve a valid
history-dependent predictor.

Physical experience is recorded as state-action-outcome transitions
\(e_{t} = (\mathbf{x}_{t},\mathbf{u}_{t},\mathbf{x}_{t + 1})\).
The internal dynamics model predicts

\begin{equation}
\widehat{\mathbf{x}}_{t + 1} =
F_{\widehat{\boldsymbol{\eta}}}(\mathbf{x}_{t},\mathbf{u}_{t},\widehat{\mathbf{z}}_{t}),\quad
\widehat{\boldsymbol{\eta}} =
(\widehat{c}_{1},\widehat{c}_{2},\widehat{k}_{w},\widehat{c}_{w},\widehat{\tau}_{a})
  \label{eq:dynamics-predictor}
\end{equation}

and the residual

\begin{equation}
\mathbf{r}_{t} = \mathbf{x}_{t + 1} - {\widehat{\mathbf{x}}}_{t + 1}
  \label{eq:residual}
\end{equation}

measures disagreement between predicted and realized motion. Accumulated
transitions update \(\widehat{\mathbf{\eta}}\) from cumulative
sufficient statistics. These identification experiments deliberately use
the correct governing equation family; they test refinement of an
inaccurate physical model, not discovery of arbitrary unknown dynamics.

\subsection{Continual route inference and matched interventions}

Candidate routes are evaluated on the current internal topology and the
lowest-cost feasible route is selected. Four information regimes isolate
distinct mechanisms. \textbf{Frozen} retains the initial map after an
environmental change. \textbf{Local} revises a gate only after it enters
the observation radius. \textbf{Oracle} receives the changed gate state
immediately and is used as a full-information spatial reference, not as
an upper bound on physical performance. The continual-memory controller
receives the same current observations as Local but can use SCM to
predict a downstream route-relevant state before direct observation when
accumulated evidence supports that inference; otherwise it falls back to
the reactive belief.

No-intervention episodes verify that the information regimes have
comparable behavior when the environment remains unchanged. In
decision-changing episodes, a transition used by the selected route
becomes invalid. In decision-preserving episodes, a structural change is
observed and incorporated into the internal model but does not alter the
currently correct decision. A separate transfer experiment withholds one
spatial transition completely from training. Position-specific and
episodic memories therefore contain no history for that location,
whereas shared SCM can exploit the relation learned elsewhere. Finally,
a matched random-change condition preserves the marginal state
frequencies while destroying the cross-transition relation.

\subsection{Evaluation and statistical analysis}

For matched navigation trials, anticipatory lead is defined as the time
at which the reactive controller first directly observes the downstream
change minus the SCM inference time. Reactive denotes the Local
controller without cross-episode memory. The representative trajectory
is visualized using a direction-coded radial distance: its magnitude is
the ball's radial distance from the maze center, while the sign records
the instantaneous circumferential direction (positive for clockwise and
negative for anticlockwise). The sign therefore encodes direction and
should not be interpreted as a negative Euclidean distance.

For the population robustness analysis, obsolete-target persistence is
analyzed primarily as a continuous paired outcome. A 0.5-s threshold is
used only as a conservative visual reference for a secondary binary
classification of substantial obsolete pursuit. The categorical result
is insensitive to the particular reference value because the two
empirical distributions are completely separated in the matched sample:
the longest SCM persistence is 0.0083 s, whereas the shortest Reactive
persistence is 2.79 s.

Random-seed or training-replicate identity is retained in paired
comparisons. Bootstrap 95\% confidence intervals quantify paired mean
differences, Wilcoxon signed-rank tests are used for paired continuous
outcomes when normality is not assumed, and exact paired tests are used for binary outcomes. The predictability sweep is also compared with the
analytical Bayes limit of the controlled change process. Additional experimental protocols, parameter settings, statistical procedures, and numerical results are provided in Appendices A and B.

\section{Results}

\subsection{Reactive adaptation remains observation-bound}

\begin{figure}[tbp]
  \centering
  \includegraphics[width=0.98\linewidth]{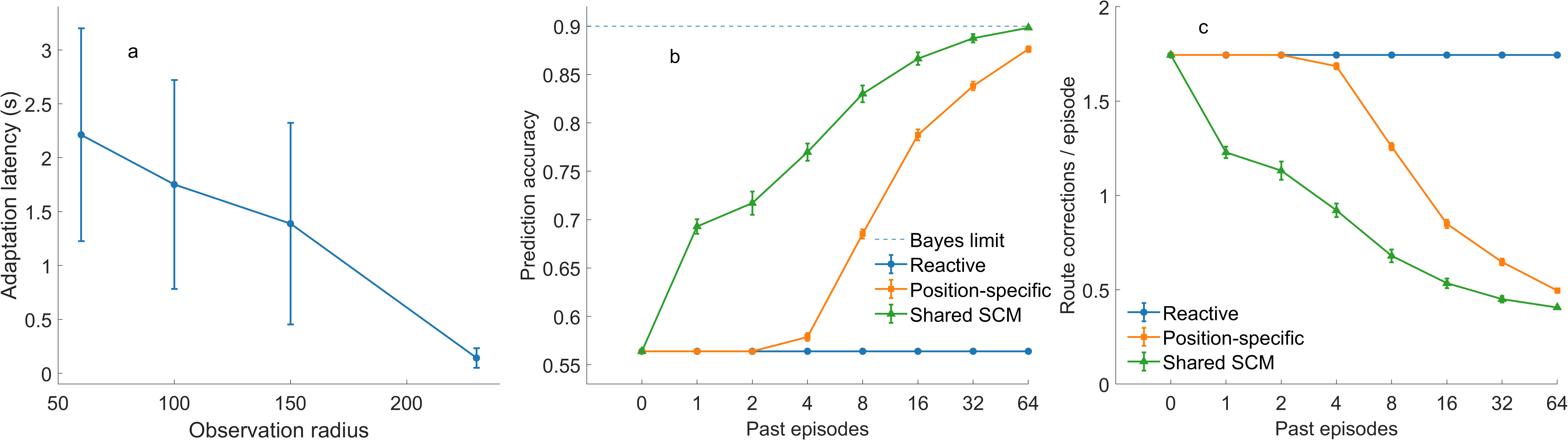}
  \caption{Reactive updating and accumulation across encounters. (a) Local adaptation accelerates as changed gates become observable sooner. (b) Shared SCM approaches the Bayes limit with accumulated episodes, while position-specific memory learns more slowly. (c) The same experience reduces later route corrections; error bars show SEM.}
  \label{fig:reactive-accumulation}
\end{figure}

When the maze remained unchanged, Frozen, Local, and Oracle agents
behaved comparably, indicating that the information regime did not
create an intrinsic baseline advantage. Closing the gate currently
supported by the route controller produced the expected separation.
Frozen continued toward the stale route and failed, whereas Local
recovered after the changed gate became observable and Oracle abandoned
the invalid route almost immediately. The latency of Local adaptation
fell monotonically as the observation radius increased, and the distance
from the event location to the changed gate strongly predicted the delay
before route revision. A correct reactive update rule can therefore
recover from change, but it remains limited by when the relevant
evidence becomes available. The dependence of reactive adaptation on
information access is shown in Figure~\ref{fig:reactive-accumulation}(a).

This provides the baseline for continual learning. Previous successful
corrections do not automatically make a later related change easier to
resolve. The following experiments test whether repeated encounters
alter the predictive state of the model and thereby reduce future
correction.

\subsection{Structured experience becomes reusable and transferable memory}

Under structured change with \(\varepsilon = 0.10\), shared SCM improved
steadily as previous encounters accumulated. Mean prediction accuracy
rose from 0.564 without usable history to 0.898 after 64 episodes, close
to the Bayes limit of 0.90. Mean route corrections fell in parallel from
1.744 to 0.406 per episode. Position-specific memory eventually
benefited from repeated exposure, but substantially more data were
required because evidence acquired at one location could not be used
immediately at another. At 64 episodes, the paired accuracy improvement
of shared SCM over the reactive condition was 0.3345 (bootstrap 95\% CI
0.3320-0.3368; Wilcoxon \(p < 2 \times 10^{-18}\)). The corresponding
prediction-accuracy and route-correction learning curves are shown in
Figures~\ref{fig:reactive-accumulation}(b) and~\ref{fig:reactive-accumulation}(c).

The withheld-transition experiment provides a stronger test of reuse.
One spatial transition was absent from all training episodes. Reactive,
episodic, and position-specific methods remained near 0.566 on this
transition, whereas shared SCM reached 0.898. The benefit therefore
cannot be attributed to retrieval of an identical route or to a
gate-specific history. What transferred was the relation learned from
other spatial contexts. The held-out comparison across shared, episodic,
position-specific, and reactive memory is shown in Figure~\ref{fig:transfer-boundary}(c).

\subsection{Predictability defines the boundary of continual memory}

\begin{figure}[tbp]
  \centering
  \includegraphics[width=0.92\linewidth]{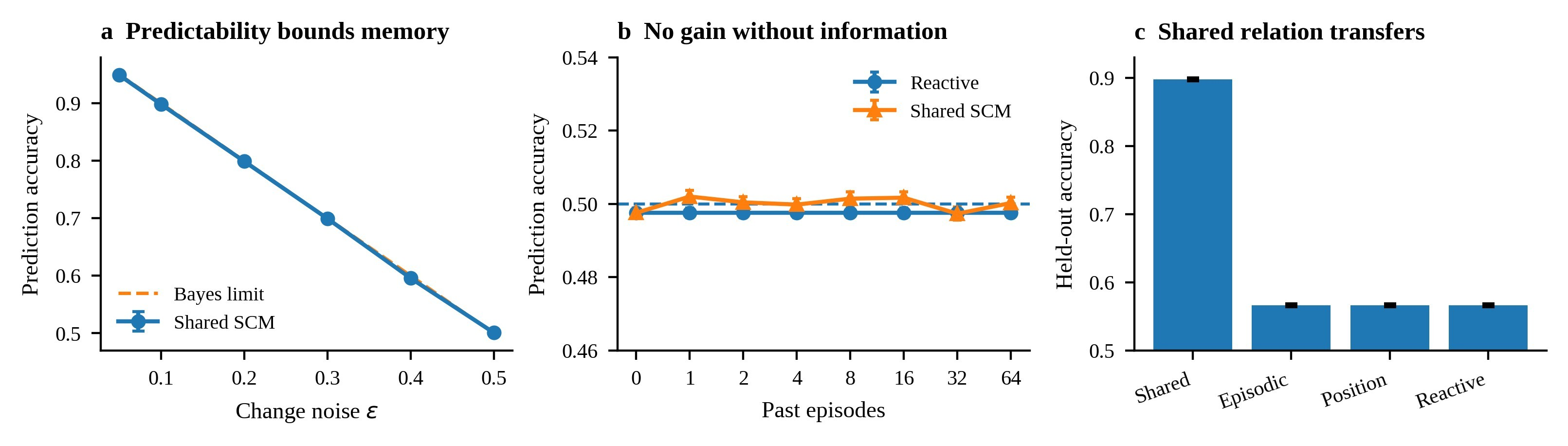}
  \caption{Transfer and the information boundary. (a) Shared SCM follows the theoretical prediction limit as environmental change becomes less predictable. (b) When past and future changes are independent, additional episodes do not improve prediction. (c) The shared relation transfers to a spatial transition excluded completely from training, unlike episodic and position-specific memories. Error bars show SEM across training replicates.}
  \label{fig:transfer-boundary}
\end{figure}

The benefit of SCM decreased systematically as the relation between
successive changes was weakened. Across \(\varepsilon = 0.05\) to
\(0.50\), empirical accuracy tracked the analytical limit
\(1 - \varepsilon\). The memory advantage also decreased with the mutual
information available in the change process. At \(\varepsilon = 0.50\),
where \(I\left( C_{k};C_{k - 1} \right) = 0\), increasing the amount of
previous experience left prediction at chance. The predictability sweep
and the zero-information learning control are shown in Figures~\ref{fig:transfer-boundary}(a)
and~\ref{fig:transfer-boundary}(b), respectively.

This control distinguishes useful accumulation from indiscriminate
history dependence. More memory is helpful only when the history
contains structure that remains informative for a later encounter. A
benchmark containing recurring change but no matched zero-information
condition may therefore make a memory mechanism appear increasingly
capable even when it has not learned a transferable relation.

\subsection{Physical experience makes continual prediction accountable to action}

\begin{figure}[tbp]
  \centering
  \includegraphics[width=0.86\linewidth]{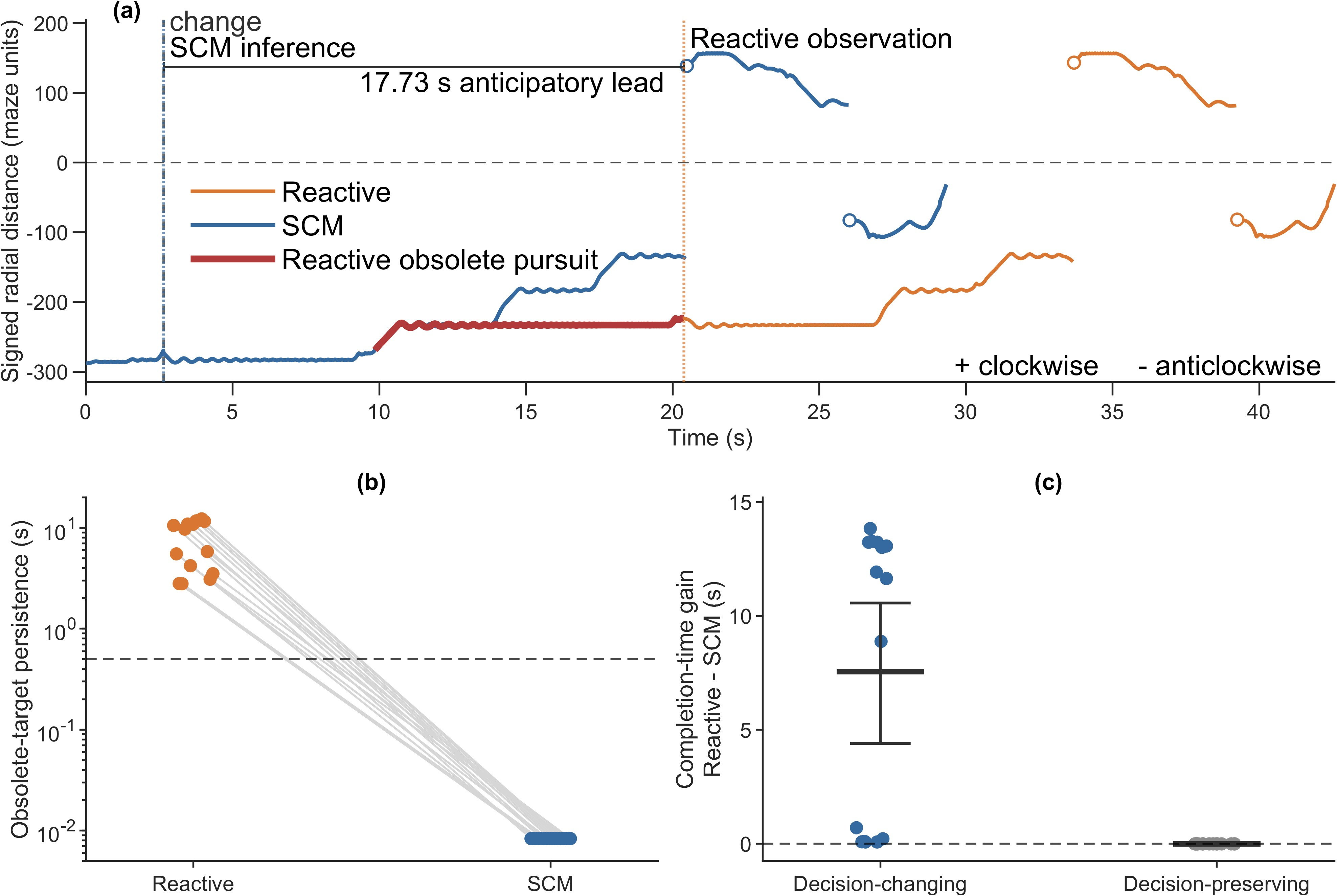}
  \caption{Predictive memory changes physical navigation before direct observation. (a) Direction-coded radial distance for representative matched seed 91000; positive and negative values denote clockwise and anticlockwise motion. The Reactive obsolete-pursuit interval is highlighted in red. Vertical markers show the structural change, SCM inference, and Reactive direct observation; the bracket marks the 17.73-s anticipatory lead. (b) Obsolete-target persistence for all 15 matched decision-changing episodes on a logarithmic axis. The distributions are completely separated: max(SCM) = 0.0083 s \textless{} min(Reactive) = 2.79 s; the 0.5-s line is a visual reference only. (c) Completion-time gain (Reactive $-$ SCM). Points are individual held-out episodes; the bar and whiskers show the mean and bootstrap 95\% confidence interval for decision-changing episodes. Mean gain is 7.56 s, while all 15 decision-preserving episodes remain exactly at zero.}
  \label{fig:physical-navigation}
\end{figure}

Continuous navigation tests whether the learned spatial relation changes
behavior selectively. In 15 held-out decision-changing episodes, SCM
reduced obsolete-target persistence from 7.806 s to one simulation step
($\Delta t=1/120$~s = 0.00833~s), a 99.89\% reduction, and shortened mean
completion time from 42.25 s to 34.69 s. The paired obsolete-target
reduction was 7.798 s (bootstrap 95\% CI 5.90--9.69 s; one-sided
Wilcoxon $p = 3.05 \times 10^{-5}$). SCM inference preceded the
time at which the reactive controller directly observed the downstream
change by 15.59 s on average. In the representative matched seed 91000,
this anticipatory lead was 17.73 s. Figure~\ref{fig:physical-navigation}(a) shows the corresponding
trajectory-level consequence: after the environmental change, Reactive
control continues along a direction-coded radial trajectory while
pursuing a now-obsolete target, whereas SCM redirects before the target
itself becomes locally observable.

The representative episode is supported by a complete matched-population
separation. Every one of the 15 Reactive episodes exhibits substantial
obsolete pursuit, whereas every SCM episode terminates obsolete pursuit
after one simulation step (Figure~\ref{fig:physical-navigation}(b)). This conclusion does not depend
on the 0.5-s visual threshold: max(SCM) = 0.0083 s is smaller than
min(Reactive) = 2.79 s, and the two-sided exact paired binary contrast
is $p = 6.10 \times 10^{-5}$. The behavioral consequence is
selective: decision-changing episodes gain 7.56 s in completion time on
average, whereas all 15 matched decision-preserving episodes have
exactly zero completion-time difference (Figure~\ref{fig:physical-navigation}(c)). Thus accumulated
spatial experience changes physical action when the previous decision is
invalidated, but does not create a general tendency to replan when the
currently correct decision remains valid. A complementary physical-model
test isolates a different source of error: with topology fixed, changing
embodied dynamics can alter execution, and cumulative
state-action-outcome updates reduce parameter error in the prescribed
equation family; at 2000 controlled experiences the largest remaining
parameter error is 4.69\% (Table~\ref{tab:appendix-checks}). The resulting evidence is
summarized in Table~\ref{tab:main-evidence}.

\begin{table}[t]
  \caption{Main evidence supporting the continual-world-model interpretation.}
  \label{tab:main-evidence}
  \centering
  \small

\begin{tabular}{p{0.18\linewidth}p{0.23\linewidth}p{0.25\linewidth}p{0.22\linewidth}}
\toprule
\textbf{Test} & \textbf{Main comparison} & \textbf{Observation} &
\textbf{Claim supported} \\
\midrule
Structured accumulation & 0 vs 64 past episodes & accuracy 0.564
\(\rightarrow\) 0.898; corrections 1.744 \(\rightarrow\) 0.406 &
experience lowers future correction \\
Held-out transfer & Shared SCM vs episodic/position/reactive & 0.898 vs
approximately 0.566 & transferable relation, not route cache \\
Zero-information control & \emph{I}=0 with increasing history & accuracy
remains approximately 0.50 & memory cannot create predictive
structure \\
Continuous navigation & memory vs reactive, decision-changing &
obsolete-target time 7.806 s \(\rightarrow\) 0.0083 s & prediction
changes action before direct observation \\
Physical accumulation & inaccurate vs continually updated dynamics &
parameter and prediction error decrease & action outcomes revise
physical prediction \\
\bottomrule
\end{tabular}
\end{table}

\section{Discussion}

\subsection{From persistent models to continual knowledge}

The experiments separate persistence from continual learning. A map,
latent state, or episode archive can remain available indefinitely
without making a later problem easier. SCM instead changes prediction on
a held-out encounter and reduces the corrections required during later
navigation. This suggests a functional criterion for continual world
models: past interaction should measurably reduce the information,
inference, or correction required in future related interaction while
preserving task performance. Prediction accuracy remains important, but
it is not sufficient by itself. Repeated success can be obtained by
solving essentially the same inference problem from current observations
on every encounter.

The continuous-navigation tests clarify a separate role of the
accumulated model. The 15 matched decision-changing episodes are
independent held-out evaluations of an already learned SCM relation;
their consistent separation therefore establishes behavioral robustness
rather than another exposure-dependent learning curve. Continual
improvement is demonstrated in Figure~\ref{fig:reactive-accumulation};
Figure~\ref{fig:transfer-boundary} establishes its information boundary
and transfer beyond trained locations. Figure~\ref{fig:physical-navigation}
addresses a different question: whether the accumulated relation changes
physical behavior before the relevant change becomes directly
observable. The representative trajectory, matched population, and
decision-preserving control show that it does so selectively, without
treating the arbitrary ordering of held-out episodes as evidence of
further within-agent learning.

The random-change condition supplies the other half of this criterion.
Nonstationarity alone does not justify continual prediction from
history. The usefulness of memory is bounded by the information that
previous states carry about later ones. A reliable continual model
should exploit repeated structure when it exists and lose that advantage
when the structure is removed. This boundary is especially important for
benchmarks in which repeated patterns may otherwise be mistaken for
evidence of general continual adaptation.

\subsection{From memories of states to models of transformations and consequences}

The transfer experiment identifies what makes the spatial memory
reusable. Episodic and position-specific memories retain where an event
occurred. Shared SCM retains a relation in how spatial structure
changes. The latter transfers because the transformation, rather than
the location, is shared. This complements incremental search and
Experience Graphs, which reuse previous search effort or successful
motion {[}17-19{]}, by treating change in the environment itself as the
object from which transferable structure can be learned.

For an embodied agent, however, transformation memory is still
incomplete if it is detached from the consequences of action.
Figure~\ref{fig:physical-navigation}
makes the first part of this coupling directly observable:
experience-derived spatial prediction changes the selected action before
local observation, and the resulting difference is expressed as a
different physical trajectory. The actuator-lag counterexample shows
that identical topology can support different outcomes under different
dynamics. A physically meaningful experience must therefore retain
state, action, and consequence in addition to spatial context. Past
outcomes constrain what should happen next, while current outcomes can
invalidate that expectation. Continual spatial world modeling then
becomes a two-way process: previous encounters provide predictive
structure, but the external world retains the authority to revise it
through the consequences of action.

\subsection{Scope and next tests}

The controlled design makes the central comparisons identifiable but
limits the claims. Spatial change is binary and deliberately structured;
a visible gate is observed without recognition error; the physical state
is low-dimensional; and parameter learning assumes the correct equation
family. The results do not establish general visual place recognition,
arbitrary dynamics discovery, or robustness to unrestricted distribution
shift. The integrated force-level navigation experiments are mechanistic
tests rather than a large-scale performance benchmark; the strongest
population-level evidence concerns cross-episode spatial-change
learning, its information boundary, held-out transfer, and matched
continuous-navigation behavior.

The immediate extension is a physical tilting maze with camera-based
ball tracking and measured actuator angles. Friction variation, contact
uncertainty, backlash, delay, and unmodeled effects would then arise
naturally rather than being prescribed. The relevant question is
unchanged: does accumulated experience reduce future correction while
prediction residuals remain able to reveal when the internal model has
become stale? The same logic can be carried to more complex navigation
and manipulation tasks, but doing so will require learned state
representations, uncertainty-aware change models, and physical
adaptation beyond a fixed parametric family.

\section{Conclusion}

Repeated interaction does not become continual world knowledge merely
because it is retained. In the dynamic maze, structured experience
became useful when it supported prediction at an unseen spatial
transition, reduced later route correction, and lost its benefit when
environmental history ceased to contain predictive information.
Physics-grounded simulation tests a second requirement: spatial
correctness alone can be insufficient for action, and state-action
outcomes provide evidence for revising the physical prediction. These
findings support a compact view of continual spatial world models as
representations that accumulate reusable structure while remaining
corrigible by the world they predict. The next test is whether the same
experience-prediction-revision cycle survives in an instrumented
physical maze where the dynamics are no longer known by construction.

\newpage
\section*{References}
{\small
[1] Sutton, R. S. (1991). Dyna, an integrated architecture for
learning, planning, and reacting. \emph{ACM SIGART Bulletin}, 2,
160-163.

[2] Deisenroth, M. P., \& Rasmussen, C. E. (2011). PILCO: A
model-based and data-efficient approach to policy search.
\emph{Proceedings of ICML}, 465-472.

[3] Ha, D., \& Schmidhuber, J. (2018). Recurrent world models
facilitate policy evolution. \emph{Advances in Neural Information
Processing Systems}, 31.

[4] Hafner, D., Lillicrap, T., Fischer, I., et al.~(2019). Learning
latent dynamics for planning from pixels. \emph{Proceedings of ICML},
2555-2565.

[5] Hafner, D., Lillicrap, T., Ba, J., \& Norouzi, M. (2020). Dream
to control: Learning behaviors by latent imagination. \emph{ICLR}.

[6] Schrittwieser, J., Antonoglou, I., Hubert, T., et al.~(2020).
Mastering Atari, Go, chess and shogi by planning with a learned model.
\emph{Nature}, 588, 604-609.

[7] Hafner, D., Pasukonis, J., Ba, J., \& Lillicrap, T. (2023).
Mastering diverse domains through world models. \emph{arXiv:2301.04104}.

[8] Tolman, E. C. (1948). Cognitive maps in rats and men.
\emph{Psychological Review}, 55, 189-208.

[9] O'Keefe, J., \& Dostrovsky, J. (1971). The hippocampus as a
spatial map: Preliminary evidence from unit activity in the
freely-moving rat. \emph{Brain Research}, 34, 171-175.

[10] Moser, E. I., Kropff, E., \& Moser, M.-B. (2008). Place cells,
grid cells, and the brain's spatial representation system. \emph{Annual
Review of Neuroscience}, 31, 69-89.

[11] Durrant-Whyte, H., \& Bailey, T. (2006). Simultaneous
localization and mapping: Part I. \emph{IEEE Robotics \& Automation
Magazine}, 13, 99-110.

[12] Thrun, S., Burgard, W., \& Fox, D. (2005). \emph{Probabilistic
Robotics}. MIT Press.

[13] Parisotto, E., \& Salakhutdinov, R. (2018). Neural Map:
Structured memory for deep reinforcement learning. \emph{ICLR}.

[14] Savinov, N., Dosovitskiy, A., \& Koltun, V. (2018).
Semi-parametric topological memory for navigation. \emph{ICLR}.

[15] Fang, K., Toshev, A., Fei-Fei, L., \& Savarese, S. (2019).
Scene Memory Transformer for embodied agents in long-horizon tasks.
\emph{CVPR}, 538-547.

[16] Chaplot, D. S., Salakhutdinov, R., Gupta, A., \& Gupta, S.
(2020). Neural Topological SLAM for visual navigation. \emph{CVPR},
12875-12884.

[17] Koenig, S., \& Likhachev, M. (2002). D* Lite. \emph{AAAI},
476-483.

[18] Koenig, S., Likhachev, M., \& Furcy, D. (2004). Lifelong
Planning A*. \emph{Artificial Intelligence}, 155, 93-146.

[19] Phillips, M., Cohen, B., Chitta, S., \& Likhachev, M. (2012).
E-Graphs: Bootstrapping planning with Experience Graphs. \emph{Robotics:
Science and Systems}.

[20] Tan, J., Zhang, T., Coumans, E., et al.~(2018). Sim-to-Real:
Learning agile locomotion for quadruped robots. \emph{Robotics: Science
and Systems}.

[21] Hwangbo, J., Lee, J., Dosovitskiy, A., et al.~(2019). Learning
agile and dynamic motor skills for legged robots. \emph{Science
Robotics}, 4, eaau5872.

[22] Sakagami, R., Lay, F. S., Doemel, A., Schuster, M. J.,
Albu-Schaeffer, A., \& Stulp, F. (2023). Robotic world models:
Conceptualization, review, and engineering best practices.
\emph{Frontiers in Robotics and AI}, 10, 1253049.

[23] Kirkpatrick, J., Pascanu, R., Rabinowitz, N., et al.~(2017).
Overcoming catastrophic forgetting in neural networks. \emph{Proceedings
of the National Academy of Sciences}, 114, 3521-3526.

[24] Parisi, G. I., Kemker, R., Part, J. L., Kanan, C., \& Wermter,
S. (2019). Continual lifelong learning with neural networks: A review.
\emph{Neural Networks}, 113, 54-71.

[25] Liu, Z., Fu, G., Du, C., Lee, W. S., \& Lin, M. (2025).
Continual reinforcement learning by planning with online world models.
\emph{arXiv:2507.09177}.
}

\appendix

\section{Experimental protocol and reproducibility}

\subsection{Structured spatial-change experiments}

The structured-memory experiments use a balanced binary change process
with \(P\left( C_{k - 1} = C_{k} \right) = 1 - \varepsilon\). The
primary learning curve uses \(\varepsilon = 0.10\). Training exposure is
varied over 0, 1, 2, 4, 8, 16, 32, and 64 episodes. Each exposure level
is evaluated across independent training replicates on fixed held-out
evaluation episodes. Reactive prediction uses no cross-episode
information; position-specific memory estimates a separate relation for
each location; episodic memory can reuse an exact stored transition;
shared SCM pools the relation across equivalent spatial positions. In
the transfer experiment, the evaluated spatial transition is excluded
from every training episode, so only a representation shared across
locations can transfer nontrivial information to that test case.

The predictability experiment holds training exposure fixed while
varying \(\varepsilon\) from 0.05 to 0.50. For the balanced binary
process, Bayes accuracy is \(1 - \varepsilon\). The random-learning
control uses \(\varepsilon = 0.50\) at all exposure levels. It preserves
the same binary state marginals while eliminating predictive dependence
between successive changes.

\subsection{Reactive and continuous-navigation experiments}

Reactive adaptation is evaluated under no intervention and current-route
closure. Frozen, Local, and Oracle use the same route controller but
differ in access to the changed topology. The observation-radius sweep
changes only the distance at which a gate becomes observable. Adaptation
latency is measured from the environmental intervention to the first
route revision consistent with the new state.

The matched continuous-navigation experiment contains 30 held-out seeds:
15 decision-changing and 15 decision-preserving episodes. The two
controller variants receive the same current observations and differ
only in whether the learned SCM relation can be used before direct
downstream observation. Obsolete-target persistence measures time spent
pursuing a transition that has already become invalid.
Decision-preserving trials serve as a specificity control: an internal
topology update occurs, but the route that should be followed does not
change.

\subsection{Physical model and identification}

The physical state transition uses a solid-sphere effective mass, linear
and quadratic drag, compliant contact, and first-order actuator
dynamics. Unit tests verify the analytical effective mass, free-roll
acceleration, and actuator step response to numerical precision. The
controlled parameter-identification experiment begins from deliberately
inaccurate estimates and updates the parameters from accumulated
actuator, drag, and contact transitions. The simulator's reference
values for this identification experiment are \(\tau_{a} = 0.18\) s,
\(c_{1} = 0.02352\) kg/s, \(c_{2} = 0.01925\) kg/m, \(k_{w} = 80\) N/m,
and \(c_{w} = 0.42\) N s/m. The physical-learning result should
therefore be interpreted as identifiability and continual refinement
within a known model family.

\subsection{Statistical reporting}

Learning curves report means across independent training replicates;
figure error bars show standard error of the mean. Paired seed identity
is preserved for continuous-navigation comparisons, and bootstrap
confidence intervals are computed from paired differences. Unless
otherwise stated, inferential tests are two-sided. The
SCM-versus-Reactive obsolete-persistence comparison uses the
pre-specified directional hypothesis that reusable structured memory
reduces obsolete action, giving a one-sided paired Wilcoxon $p = 3.05 \times 10^{-5}$ for 15 matched pairs. The secondary
substantial-pursuit classification (\textgreater0.5 s) yields 15/15
Reactive versus 0/15 SCM and a two-sided exact paired binary $p = 6.10 \times 10^{-5}$. Because every SCM value is 0.0083 s and every
Reactive value is at least 2.79 s, this binary conclusion is invariant
for any threshold strictly between those values.

\section{Additional numerical results}

\begin{table}[t]
  \caption{Selected quantitative checks and boundary conditions.}
  \label{tab:appendix-checks}
  \centering
  \small

\begin{tabular}{p{0.27\linewidth}p{0.32\linewidth}p{0.31\linewidth}}
\toprule
\textbf{Quantity} & \textbf{Result} & \textbf{Role in the study} \\
\midrule
Effective mass check & expected 0.056 kg; measured 0.056 kg & verifies
rolling-sphere reduction \\
Observation-radius effect & adaptation latency decreases monotonically
with radius & establishes information-access limit \\
Distance-latency association & Spearman $\rho=0.812$, $p=4.98 \times 10^{-8}$ & links reactive delay to
information arrival \\
Structured SCM at 64 episodes & accuracy 0.898; Bayes limit 0.900 &
shows near-limit use of repeated structure \\
Random change at 64 episodes & accuracy approximately 0.50 &
zero-information negative control \\
Held-out shared SCM & accuracy 0.898 vs approximately 0.566 baselines &
tests transfer beyond stored location \\
Decision-preserving navigation & zero completion-time difference for all
matched seeds & tests behavioral specificity \\
Physical identification at 2000 experiences & maximum parameter error
4.69\% & bounds same-family model refinement \\
\bottomrule
\end{tabular}
\end{table}

The route/gate-identifier memory tested during method development is not
included as a positive result because its learned identity statistics
did not improve held-out physical navigation reliably. This negative
control reinforces the distinction between retaining an
identifier-specific history and learning a relation that transfers
across encounters.

\end{document}